# Quantitative causality, causality-guided scientific discovery, and causal machine learning

X. San Liang[1,2], Dake Chen[1], Renhe Zhang[2]

[1] Division of Frontier Research, Southern Marine Laboratory,
and School of Artificial Intelligence, Sun Yat-Sen University, Zhuhai, China

[2] Department of Atmospheric and Oceanic Sciences,
Fudan University, Shanghai, China

**Abstract**

It has been said, arguably, that causality analysis should pave a promising way to interpretable deep learning and generalization. Incorporation of causality into artificial intelligence (AI) algorithms, however, is challenged with its vagueness, non-quantitiveness, computational inefficiency, etc. During the past 18 years, these challenges have been essentially resolved, with the establishment of a rigorous formalism of causality analysis initially motivated from atmospheric predictability. This not only opens a new field in the atmosphere-ocean science, namely, information flow, but also has led to scientific discoveries in other disciplines, such as quantum mechanics, neuroscience, financial economics, etc., through various applications. This note provides a brief review of the decade-long effort, including a list of major theoretical results, a sketch of the causal deep learning framework, and some representative real-world applications in geoscience pertaining to this journal, such as those on the anthropogenic cause of global warming, the decadal prediction of El Niño Modoki, the forecasting of an extreme drought in China, among others.

## 1. Introduction

Causality analysis is a fundamental problem in scientific research, as commented by Einstein in 1953 in response to a question on the *status quo* of science in China at that time (*cf.* the historical record in Hu, 2005). The recent rush in artificial intelligence (AI) has stimulated enormous interest in causal inference, partly due to the realization that it may take the field to the next level to approach human intelligence (see Pearl, 2018; Bengio, 2019; Schölkopf, 2022). In the fields pertaining to this journal, assessment of the cause-effect relations between dynamic events makes a natural objective for the corresponding researches.

During the past 18 years, causality analysis in a quantitative sense has been developed independently in physics from first principles. This is quite different from the various formalisms such as the classical ones in statistics, e.g., Granger (1969). It is the first formalism established on a rigorous footing within the framework of dynamical systems, which yields an explicit solution in closed form, allowing for quantifying and normalizing with ease the causality between dynamical events. Originally born from atmospheric predictability studies, it is hence of special use to the earth system science and, particularly, the atmosphere-ocean-climate science. In the following we will give a brief introduction of the theory and the resulting methodology, with a focus on a concise formula for the sake of practical usage. We then show, with the aid of this methodology, how a variety of scientific discoveries can be easily made. This is followed by an introduction of the recent causal AI algorithm development, and its application to a long-term prediction of El Niño Modoki, and the forecasting of an extreme drought in China.

## 2. A brief stroll through part of the theory of quantitative causality analysis

Ever since Granger's seminal work (Granger, 1969), data-based causal inference has been conventionally investigated as a subject in statistics or engineering science. On the other hand, Liang (2016) argued that causality in the Newtonian sense is actually "a real physical notion that can be derived *ab initio*." This line of work can be traced back to Liang and Kleeman (2005), where a discovery about the information flow with two-dimensional deterministic systems is presented. A comprehensive study with generic systems has been fulfilled recently, with explicit formulas attained in closed forms; see Liang (2008) and Liang (2016). These formulas have been validated with benchmark systems such as baker transformation, Hénon map, Kaplan-Yorke map, Rössler system, to name a few. They have also been applied to real world problems in the diverse disciplines such as climate science, meteorology, turbulence, neuroscience, financial economics, quantum mechanics, etc. The following is a brief introduction of the theory, originally motivated by the predictabilty study in atmosphere-ocean science and later on formulated within a more generic framework, i.e., the framework of stochastic dynamical systems.

To illustrate, consider a *d*-dimensional continuous-time stochastic system (systems with discrete-time mappings also available; see Liang, 2016):

$$\frac{d\boldsymbol{X}}{dt} = \boldsymbol{F}(\boldsymbol{X},t) + \boldsymbol{B}(\boldsymbol{X},t)\dot{\boldsymbol{W}}, \tag{1}$$

where $\boldsymbol{X} = (X_1, X_2 \ldots, X_d)^T$ is a *d*-dimensional vector of state variables, where $\boldsymbol{F} = (F_1, F_2 \ldots, F_d)^T$ may be arbitrary nonlinear functions of $\boldsymbol{X}$ and $t$, $\dot{\boldsymbol{W}}$ is a vector of white noise, and $\boldsymbol{B} = (b_{ij})$ is the matrix of stochastic perturbation amplitudes. Here $\boldsymbol{F}$ and $\boldsymbol{B}$ are both assumed to be differentiable with respect to $\boldsymbol{X}$ and $t$. For deterministic systems such as those in meteorology and oceanography, $\boldsymbol{B}$ is zero. Liang and Kleeman (2005) defined the rate of information flow/information transfer, or simply *information flow*, from a component $X_j$ to another component $X_i$, as the contribution of entropy from $X_j$ per unit time in increasing the marginal entropy of $X_i$. Hereafter by "entropy" we mean Shannon entropy, although other types of entropy have also been explored, such as Kullback–Leiber divergence (Liang, 2018), Von Neumann entropy (Yi and Bose, 2022), etc. Liang (2016) proved, using the technique of Frobenius-Perron operator, that the rate of information flowing from $X_j$ to $X_i$ (in nats per unit time), denoted $T_{j \to i}$, is

$$T_{j \to i} = -E\left(\frac{1}{\rho_i}\int_{\mathbb{R}^{d-2}} \frac{\partial F_i \rho_{\slashed{j}}}{\partial x_i} d\boldsymbol{x}_{\slashed{\slashed{j}}}\right) + \frac{1}{2}E\left(\frac{1}{\rho_i}\int_{\mathbb{R}^{d-2}} \frac{\partial^2 (g_{ii}\rho_{\slashed{j}})}{\partial x_i^2} d\boldsymbol{x}_{\slashed{\slashed{j}}}\right), \tag{2}$$

where $d\mathbf{x}_{\hat{i}\hat{j}}$ signifies $dx_1 dx_2 \ldots dx_{i-1} dx_{i+1} \ldots dx_{j-1} dx_{j+1} \ldots dx_d$, $E$ stands for mathematical expectation, $g_{ii} = \sum_{k=1}^{d} b_{ik} b_{ik}$, $\rho_i = \rho_i(x_i)$ is the marginal probability density function (pdf) of $X_i$, and $\rho_{\hat{i}} = \int_{\mathbb{R}} \rho(\mathbf{x}) dx_{\hat{i}}$.

Equation (2) has a nice property, which forms the basis of the information flow-based causality analysis (Liang 2008), and has been referred to as "principle of nil causality." It reads that

*If the evolution of $X_i$ does not depend on $X_j$, then $T_{j \to i} = 0$.*

Based on this property, the algorithm for the information flow-based causal inference is as follows: If $T_{j \to i} = 0$, then $X_j$ not causal to $X_i$; otherwise it is causal, and the absolute value measures the magnitude of the causality from $X_j$ to $X_i$.

Another property regards the invariance upon coordinate transformation, indicating that the obtained information flow is an intrinsic property in nature; see Liang (2018). As shown in Liang (2021) (and other publications), this is very important in causal graph reconstruction. It together with the principle of nil causality makes it promising toward a solution of the problem of latent confounding.

For linear systems, i.e., when $\mathbf{F}(\mathbf{X}) = \mathbf{f} + \mathbf{A}\mathbf{X}$, and when $\mathbf{B}$ is constant, then

$$T_{j \to i} = a_{ij} \frac{\sigma_{ij}}{\sigma_{ii}}, \qquad (3)$$

where $a_{ij}$ is the $(i,j)^{th}$ entry *of* $\mathbf{A}$ and $\sigma_{ij}$ the population covariance between $X_i$ and $X_j$. Notice if $X_i$ and $X_j$ are not correlated, then $\sigma_{ij} = 0$, which yields a zero causality: $T_{j \to i} = 0$. But conversely it is not true. We hence have the following corollary:

*In the linear sense, causation implies correlation, but not vice versa.*

In an explicit formmula, this corollary expresses the long-standing debate on causation vs. correlation ever since George Berkeley (1710).

In the case with only $d$ time series $X_1, X_2, \ldots, X_d$, under the assumption of a linear model with additive and independent noises, the maximum likelihood estimator (MLE) of (2) for $T_{2 \to 1}$ is*:*

$$\widehat{T}_{2 \to 1} = \frac{1}{\det \mathbf{C}} \cdot \sum_{j=1}^{d} \Delta_{2j} C_{j,d1} \cdot \frac{C_{12}}{C_{11}} \qquad (4)$$

where $C_{ij}$ is the sample covariance between $X_i$ and $X_j$, $\Delta_{ij}$ the cofactors of the matrix $\mathbf{C} = (C_{ij})$, and $C_{i,dj}$ the sample covariance between $X_i$ and a series derived from $X_j$: $\dot{X}_{j,n} = (X_{j,n+k} - X_{j,n})/(k\Delta t)$, with $k \geq 1$ some integer.

Eq. (4) is rather concise in form, involving only the common statistics, i.e., sample covariances. The transparent formula makes causality analysis, which otherwise would be complicated, very easy and computationally efficient. Note, however, that Eq. (4) cannot replace (2); it is just an estimator (MLE) of the latter. One needs to test the statistical significance before making a causal inference based on the estimator $\widehat{T}_{2 \to 1}$.

If what are given are not time series, but independent, identically distributed (i.i.d.) panel data, it has been shown that $\widehat{T}_{2 \to 1}$ has the same form as (4); see Rong and Liang (2021).

Besides the information flow between two components, say $X_1$ and $X_2$, it is also possible to estimate the influence of one component, say $X_1$, on itself. Following the convention since Liang and Kleeman (2005), write it as $dH_1^*/dt$. Then its MLE is

$$\frac{\widehat{dH_1^*}}{dt} = \frac{1}{\det \mathbf{C}} \cdot \sum_{j=1}^{d} \Delta_{1j} C_{j,d1} \qquad (5)$$

This result, first obtained in Liang (2014), provides an efficient approach to identifying self loops in a causal graph (cf. Hyttinen et al. 2012), which has been a challenging issue.

If what we want to know is the causal relation between two subsystems/subnetworks, rather than two individual components/nodes, the information flow can also be derived, in a way as above, and estimated in the maximum likelihood sense. The results are referred to Liang (2022).

Statistical significance tests can be performed for the estimators. This is done with the aid of a Fisher information matrix. See Liang (2014) and Liang (2021) for details.

Causality in this sense can be normalized in order to reveal the relative importance of a causal relation. See Liang (2015) for details.

### 3. Causality-guided scientific discoveries

The above rigorous formalism has been successfully put to application in many real world problems, and has led to important scientific discoveries which would otherwise be difficult, if not impossible, to fulfill. These include, on an incomplete list, those in the fields of global climate change (Stips et al. 2016), near-wall turbulence (Liang and Lozano-Durán 2016), atmosphere-ocean interaction (Docquier et al., 2023), Arctic rapid warming (Docquier et al, 2022), financial time economics (e.g., Lu et al., 2022; 2023), soil moisture-precipitation interaction (Hagan et al. 2018), climate network (Vannitsem and Liang, 2022), brain disease diagnosis (Hristopulos et al., 2019; Cong et al., 2023), El Niño-Indian Ocean Dipole connection (Liang et al., 2014), quantum information (Yi and Bose, 2022), to name a few. Among these we want to particularly mention the study by Stips et al. (2016) on greenhouse gases *vs.* global mean surface temperature anomaly (Fig. 1a). They found that $CO_2$ emission does drive the recent global warming during the past century, and the causal relation is one-way; Fig. 1a shows the global causal pattern. However, on a time scale of 1000 years or up, this one-way causality is completely reversed, becoming a causality from air temperature to carbon dioxide concentration. In other words, on the paleoclimate scale, it is global warming that drives the $CO_2$ emission. This interesting result is consistent with that inferred from the recent ice-core data from Antarctica.

Another discovery pertaining to the scope of this journal is the linkage of South China Sea to the Pacific-North American (PNA) teleconnection pattern. As shown in Fig. 1b, a direct application of (4) to the PNA index series and the series of the sea surface temperature (SST) reveals a clear causal pattern in the Pacific similar to the canonical El Niño, plus another center with equal causal strength but limited within South China Sea. The former verifies the well-known conclusions in previous studies that El Niño is the main drive of PNA. The latter, however, is a completely new one. This is counterintuitive, considering that South China Sea is just a margina sea with limited size, while PNA is a Northern Hemisphere climate mode that influences the North American weather. This remarkable result, which proves true later on (Zhang and Liang 2022), is just a straightforward application of the aforementioned Eq. (4).

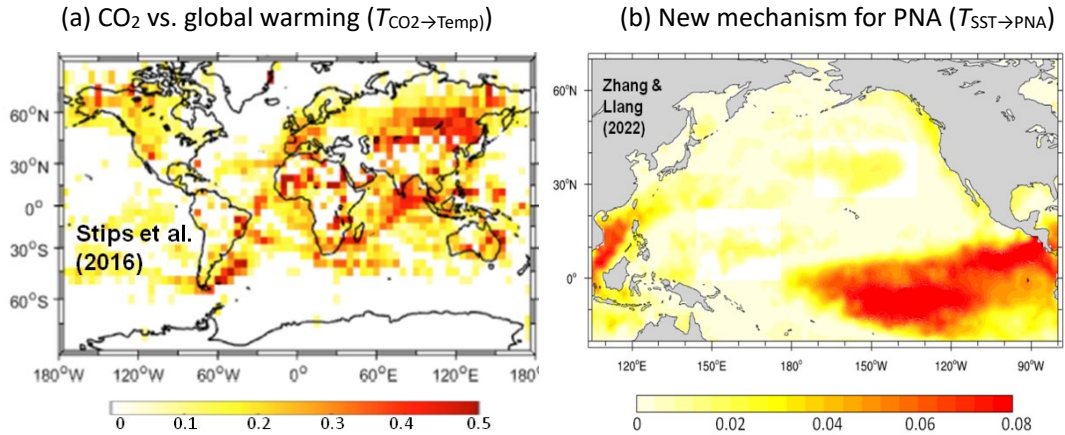

*Figure 1. (a) The causal structure between the global radiative forcing and the annual global mean surface temperature anomaly (GMTA): The information flow (in nats/yr) from $CO_2$ concentration to GMTA (from Fig. 3b of Stips et al., 2016).*

*(b) The informtion flow (in nats/mon) from sea surface temperature (SST) to the Pacific-North American teleconnection pattern (PNA) index. Note the hotspot in South China Sea, which tells a new mechanism of PNA generation (redrawn from Zhang and Liang, 2021, Fig. 1).*

## 4. Causal AI and interpretable machine learning

As mentioned in the beginning, the current impetus in causality analysis is mainly driven by the AI rush. Incorporating information flow into machine learning to build interpretable AI algorithms is hence of great interest. This is not only due to the mathematical rigor and physical interpretability in the formalism, but also due to the remarkable computing performance: on a laptop (DELL XPS MT530), for a network with 30 time series, it takes less than 1 second to fulfill the computation of the $30 \times 29 = 870$ causal relations, while using the matlab function gctest (Granger causality testing) the time needed for the computation will be more than 17 days. Efforts in this regard include predictor choosing or sparsity identification (e.g., Bai et al., 2018; Liang et al., 2021), and causal neural network construction in deep learning. The former can be simply put as the application of causality analysis prior to deep learning. The latter is more essential, requiring the building of new algorithms. The first algorithm is built by Tyrovolas et al. (2023), who have successfully embedded the Liang-Kleeman information flow analysis into the Fuzzy Cognitive Maps (FCMs). Fig. 2 is a schematic of the architecture as proposed by Tyrovolas et al. (2023). In this way the spurious correlations in the neural network are effectively removed, and explanatory power enhanced and prediction/generalization capability improved. Their numerical experiments demonstrate that the new algorithm outperforms all the state-of-the-art FCM-based models in terms of interpretability.

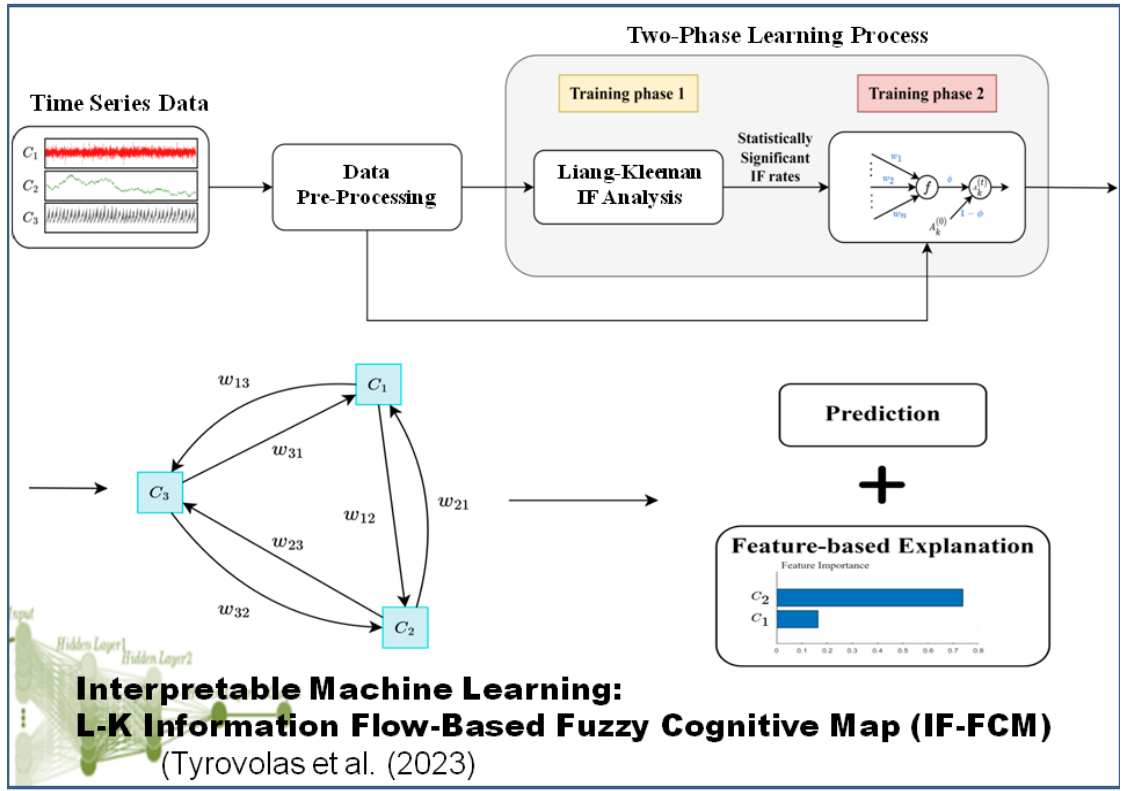

*Figure 2. The architecture of a causal AI algorithm, information flow-based fuzzy cognitive map (IF-FCM), as proposed by Tyrovolas et al. (2023)（credit: M. Tyrovolas）.*

### 5. Causal AI-based atmosphere-ocean-climate predictions

A recent application of the causal AI prediction is about the monthly precipitation over China. In 2022, the middle and lower reaches of the Yangtze River experienced an extreme drought in summer and autumn. Poyang Lake, the largest freshwater lake in China, turned into a prairie, as shown in the inserted subfigure in Fig. 3a. Equipped with the above causal AI technique, a team participated in a National Contest on monthly precipitation prediction (at a lead time of one month), and rather accurately predicted the drought in the Yangtze River reaches from July through October, 2022. Shown in Fig. 3a is the prediction for September (ratio of precipitation anomaly to climatology). Particularly, the severe situation around Poyang Lake is clearly seen.

Another major application is the one with El Niño Modoki, or Central-Pacific type El Niño. El Niño is a climate mode that has been linked to many hazards globewide, e.g., flooding, drought, wild fires, heat waves, etc. Its accurate forecasting is of great importance to many sectors of our society such as agriculture, energy, hydrology, to name several. With its societal importance and the elegant setting, El Niño prediction has become a testbed for AI algorithms.

Currently the wisdom for El Niño prediction is that it may be predictable at a lead time of 1-2 years. However, there still exists much uncertainty; an example is the 2014-16 "Monster El Niño," almost all projections fell off the mark. Among the El Niño varieties, it is believed that El Niño Modoki is particularly difficult to predict. A striking breakthrough has just been made. Liang et al. (2021) took advantage of the quantitative nature of the above information flow-based causality analysis, and identified a delayed causal pattern, i.e., the structure of the information flow from the solar activity to the sea surface temperature, very similar to the El Niño Modoki mode. They hence conjectured that, based on the series of sunspot numbers, El Niño Modoki should be predictable. This is indeed the case, and, remarkably, the prediction can be at a lead time of as long as 10 years or up (Fig. 3a). This remarkable progress, among others, is a result of the rigorously formulated quantitative causality analysis.

(a) Prediction of the 2022 severe drought in China

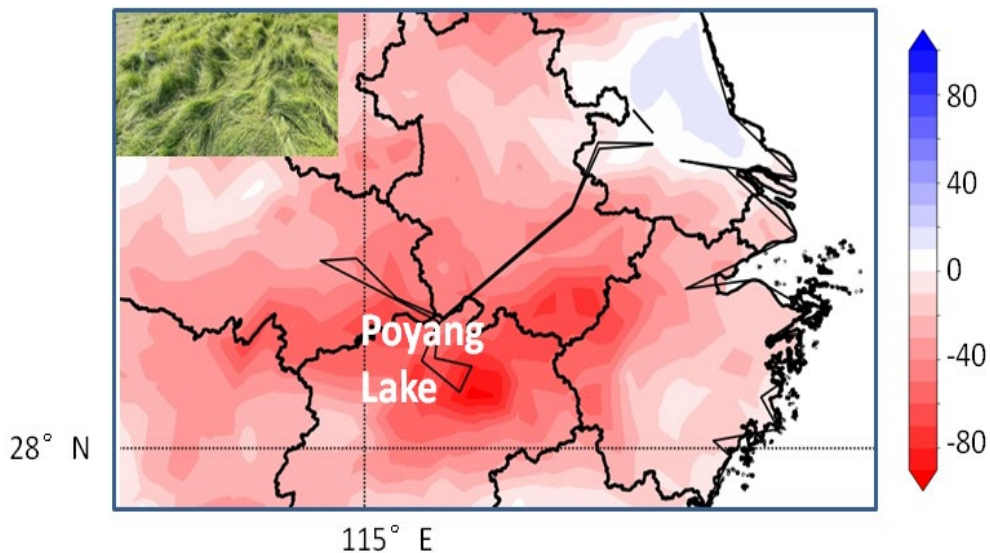

(b) Decadal prediction of El Niño Modoki

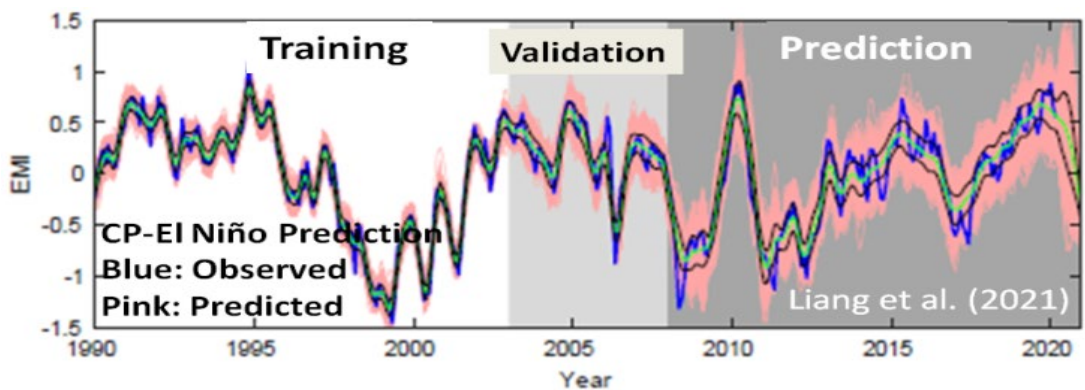

*Figure 3. (a) In the 2022 China National Contest on monthly precipitation prediction, a team equipped with the causal AI technique successfully predicted the extreme drought in the middle and lower reaches of Yangtze River. Here the colorbar indicates the ratio of precipitation anomaly (from climatology) to climatology (in percentage). Inserted is a picture of the largest freshwater lake in China in that extreme drought situation, which looks like the prairie (redrawn from Ma et al., 2022).*

*(b). El Niño prediction has become a benchmark problem for the testing of machine learning algorithms. The present wisdom is that El Niño may be predicted at a lead time of 1-2 years. Shown here are 1000 predictions (pink) of the El Niño Modoki index (EMI) as described in the text. Overlaid are the observed EMI (blue), the mean of the realizations (cyan). The light shading marks the period for validation, while the darker shading marks the prediction period (from Liang et al. 2021, Fig. 5).*

# 6. Concluding remarks

Assessment of the cause-effect relation between dynamic events is a major objective of scientific research. Recently there has been a surge of interest in data-based causal inference, echoing the arrival of the era of big data. Among the efforts is the development of a rigorous formalism of causality analysis from atmosphere-ocean science. This not only opens a new field in the atmosphere-ocean science, namely, information flow, but also has led to scientific discoveries in the diverse disciplines such as quantum mechanics, neuroscience, financial economics, etc. Moreover, its quantitativity in nature, rigor in physics, and efficiency in computation have revealed to us a promising approach to interpretable deep learning and generalization. Indeed, this has already led to the success in some notoriously difficult real-world problems. We are expecting more realistic applications in the earth system science and, particularly, in climate projection and weather forecasting.

**Acknowledgments**
This research was supported by National Science Foundation of China (Grant #42230105).

**Author Contributions**: Conceptualization, X.S.L., D.C. and R.Z.; writing, X.S.L.; validation, X.S.L., D.C. and R.Z. All authors have read and agreed to the published version of the manuscript.

**Data Availability**: Not Applicable

**Conflict of Interest**：  The authors declare no competing interests